\documentclass{article}
\usepackage{ijcai25}

\usepackage{times}
\usepackage{soul}
\usepackage{url}
\usepackage[hidelinks]{hyperref}
\usepackage[utf8]{inputenc}
\usepackage[small]{caption}
\usepackage{graphicx}
\usepackage{amsmath}
\usepackage{amsthm}
\usepackage{booktabs}
\usepackage{algorithm}
\usepackage{algorithmic}
\usepackage[switch]{lineno}
\usepackage{xcolor}
\usepackage{amssymb}
\usepackage{bbding}
\usepackage{multirow}
\usepackage{pifont}%
\usepackage{fancyhdr}

\title{Category-aware EEG image generation based on wavelet transform and contrast semantic loss\footnote{This paper has been accepted by IJCAI-2025}}

\author{
Enshang Zhang$^1$$^,$$^*$
\and
Zhicheng Zhang$^2$$^,$$^3$$^,$$^*$\and
Takashi Hanakawa$^1$\\
\affiliations
$^1$Department of Integrated Neuroanatomy and Neuroimaging, Kyoto University Graduate School of Medicine, Kyoto, Japan.\\
$^2$JancsiLab, JancsiTech, Hongkong, China\\
$^3$Sino-Finland Joint AI Laboratory for Child Health of Zhejiang Province, Hangzhou, China\\
$^*$ Corresponding Author\\
\emails
zhang.enshang.58p@st.kyoto-u.ac.jp,
zhangzhicheng13@mails.ucas.edu.cn,
hanakawa.takashi.2s@kyoto-u.ac.jp
}

\begin{document}

\maketitle

\begin{abstract}
Reconstructing visual stimuli from EEG signals is a crucial step in realizing brain-computer interfaces. In this paper, we propose a transformer-based EEG signal encoder integrating the Discrete Wavelet Transform (DWT) and the gating mechanism. Guided by the feature alignment and category-aware fusion losses, this encoder is used to extract features related to visual stimuli from EEG signals. Subsequently, with the aid of a pre-trained diffusion model, these features are reconstructed into visual stimuli. To verify the effectiveness of the model, we conducted EEG-to-image generation and classification tasks using the THINGS-EEG dataset. To address the limitations of quantitative analysis at the semantic level, we combined WordNet-based classification and semantic similarity metrics to propose a novel semantic-based score, emphasizing the ability of our model to transfer neural activities into visual representations. Experimental results show that our model significantly improves semantic alignment and classification accuracy, which achieves a maximum single-subject accuracy of 43\%, outperforming other state-of-the-art methods. The source code and supplementary material is available at \url{https://github.com/zes0v0inn/DWT_EEG_Reconstruction/tree/main}
\end{abstract}

\section{Introduction}

In recent years, the reconstruction of visual stimuli from electroencephalogram (EEG) has emerged as a highly promising research area within the domain of brain-computer interface (BCI), by extracting visually relevant features from EEG and ultimately reconstructing visual stimuli~\cite{bai2023dreamdiffusion,kavasidis2017brain2image}.
This technology has the ability to convert neural signals into images, thereby establishing a crucial link between brain activities and the external world and deepening our comprehension of the intricate relationship between brain activities and perception.
It holds immense potential, especially for individuals with severe disabilities. 
By enabling them to convey their thoughts and intentions through visual representations, it has the potential to revolutionize assistive communication methods.

With the powerful image-generation capabilities of generative networks, such as adversarial generative networks (GAN)~\cite{goodfellow2014generative}, variational auto-encoder networks (VAE)~\cite{kingma2013auto}, and denoising diffusion probabilistic models (DDPM)~\cite{zhang2025texture,ho2020denoising}, which make it possible to perceive brain visual stimuli from EEG and reconstruct visual stimuli. 
However, existing related methods face substantial challenges.
First, reconstructing visual stimuli from EEG at the pixel level is difficult and unnecessary. 
To this end, reconstructing semantically consistent images is of significant importance in BCI, leading to the inability to utilize traditional objective image evaluation indicators, such as structural similarity (SSIM). 
Therefore, the need arises for quantitatively evaluating the quality of visual stimuli to be reconstructed using an objective semantic-based score.
In addition, EEG are time-series and noisy, which complicates the extraction of visually relevant features from them. 
Moreover, the training of advanced models, such as  DreamDiffusion~\cite{bai2023dreamdiffusion} and MinD-vis~\cite{chen2023seeing}, requires an excessive amount of computational resources, severely restricting their widespread application.

To process noisy time-series EEG signals efficiently, the integration of traditional signal analysis methods, such as the discrete wavelet transform (DWT)~\cite{chen2017high}, into deep-learning modules has been employed~\cite{zeng2024wet}.  
By incorporating DWT-based modules into EEG encoder, we can leverage both the spatial and frequency characteristics of EEG, enhancing the feature-extraction process. 
In addition, with the continuous evolution of deep-learning models, the gate mechanisms in Mamba~\cite{gu2023mamba} have been proven effective in neural networks for selectively controlling information flow.

Motivated by the success of gated attention mechanisms in Mamba and DWT in signal processing, in this work, we integrated a DWT module with the gated attention mechanism within a well-designed EEG encoder, extracting meaningful features from EEG by effectively capturing both spatial and frequency-domain information while selectively focusing on relevant features.
To reconstruct semantically consistent images without pixel-level ground truth, a category-aware clustering loss is utilized to cluster samples of the same category and separate different categories in a high-dimensional space, thereby improving the reconstruction accuracy and zero-shot generalization ability of the model.
After the EEG feature extraction, we employ a pre-trained diffusion-based model to generate high-quality images. To ensure that the reconstructed images accurately reflect the semantic meaning of the input EEG, we incorporate an image classification model as a performance validation step. Subsequently, we propose a related evaluation metric of semantic-based score, enabling a quantitative assessment of the semantic consistency between the input EEG and the reconstructed images.
The novelty of this work are three-folds:
\begin{enumerate}
\item To the best of our knowledge, it is the first time to use category-aware clustering loss to better extract the EEG features associated with visual stimuli, with the assistance of the CLIP loss~\cite{radford2021learning} to align image and EEG features.
\item To evaluate the performance of reconstructing visual stimuli from EEG, we creatively propose an evaluation metric of semantic-based score based on the pre-trained classification model from the open-source community, allowing for quantitatively analysis of various methods' generation effects.
\item In view of the time-series nature and high noise level of EEG, we combine DWT and the gated attention mechanism to design a specialized EEG encoder. It achieves high performance in EEG classification and supports downstream image generation tasks.

\end{enumerate}

\section{Related Work}

Reconstructing visual stimuli from brain signals, such as functional magnetic resonance imaging(fMRI) and EEG, has achieved remarkable results in previous studies. For instance, conditional Generative Adversarial Networks and VAEs have been applied to encode fMRI signals and further reconstruct the visual stimuli from the features of brain signals. 
Studies such as Shen \textit{et al.}~\cite{shen2019deep} and Takagi \textit{et al.}~\cite{takagi2023high} demonstrated that fMRI data could be translated into semantically correct and high-quality images. In addition, EEG-based reconstruction algorithms, despite the challenges of noise and lower spatial resolution, have also exhibited promising results. Techniques integrating convolutional neural networks with GANs, as in~\cite{song2021common,yang2021survey}, have enabled the generation of visual representations corresponding to real-time brain activity. 
To simplify the application of BCI and make the technology more low-cost and convenient for use, EEG is a more practical signal entry than fMRI.
Recent outstanding results, such as DreamDiffusion~\cite{bai2023dreamdiffusion} and ATM-S~\cite{li2024visual}, have shown that, with the assistance of the powerful image generation ability of diffusion model, deep learning models can reconstruct visual stimuli from EEG, capturing some semantic information embedded in neural activity. 
Moreover, multi-modal approaches combining EEG and fMRI have enhanced image quality and robustness by leveraging complementary features from both modalities. 
These advancements emphasize the growing potential of brain-signal-based image generation for applications. 

\section{Methods}

\subsection{Overall Architecture}
\begin{figure}[htbp]
\centering
    \includegraphics[width=0.48\textwidth]{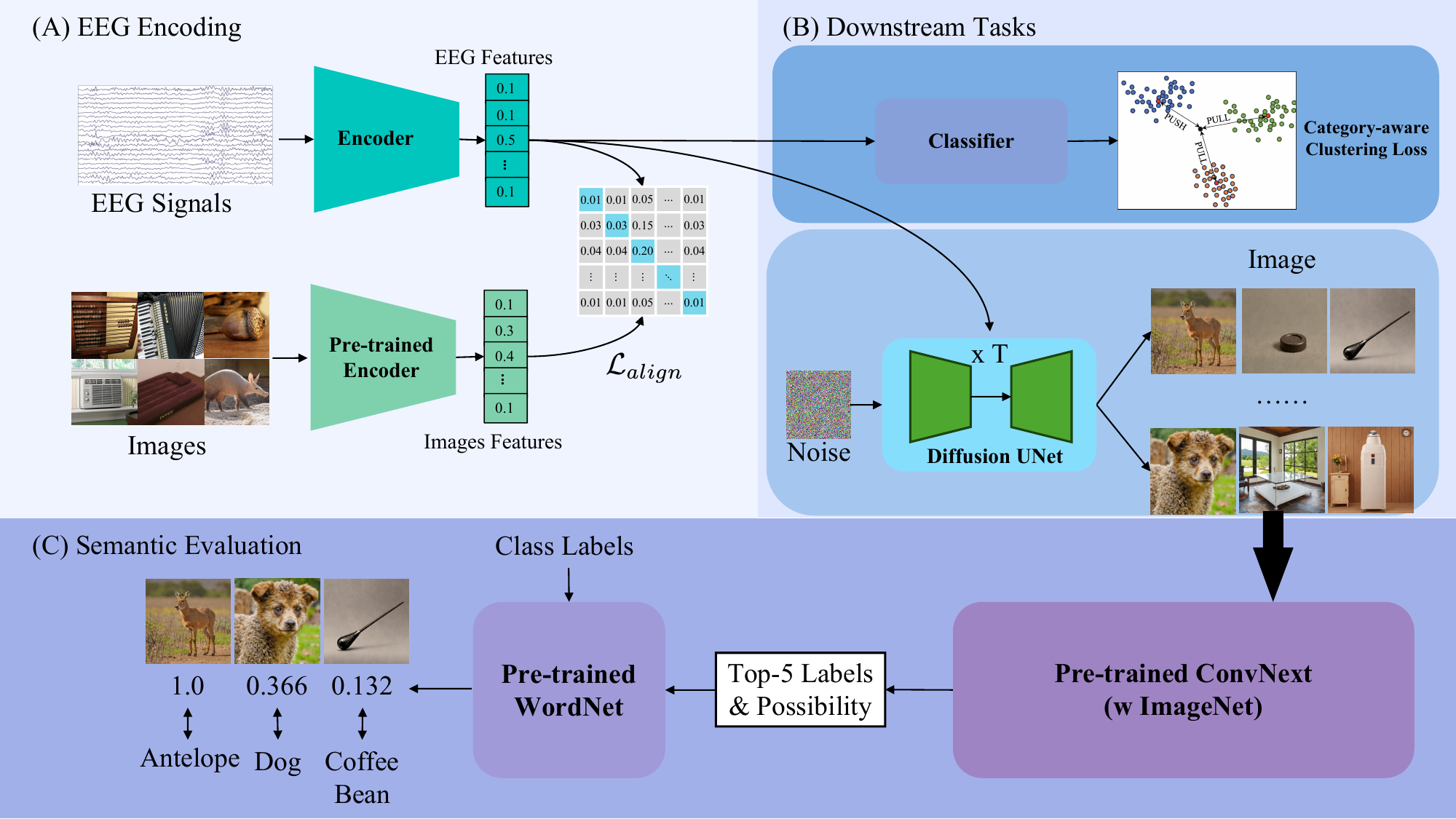}
    \caption{The overall flowchart of the proposed model. The whole model can be distinguished into three main parts including EEG embedding part (A), downstream part (B) and the semantic evaluation part (C).  }
    \label{figure2}
\end{figure}

\begin{figure}[hbp]
\centering
    \includegraphics[width=0.48\textwidth]{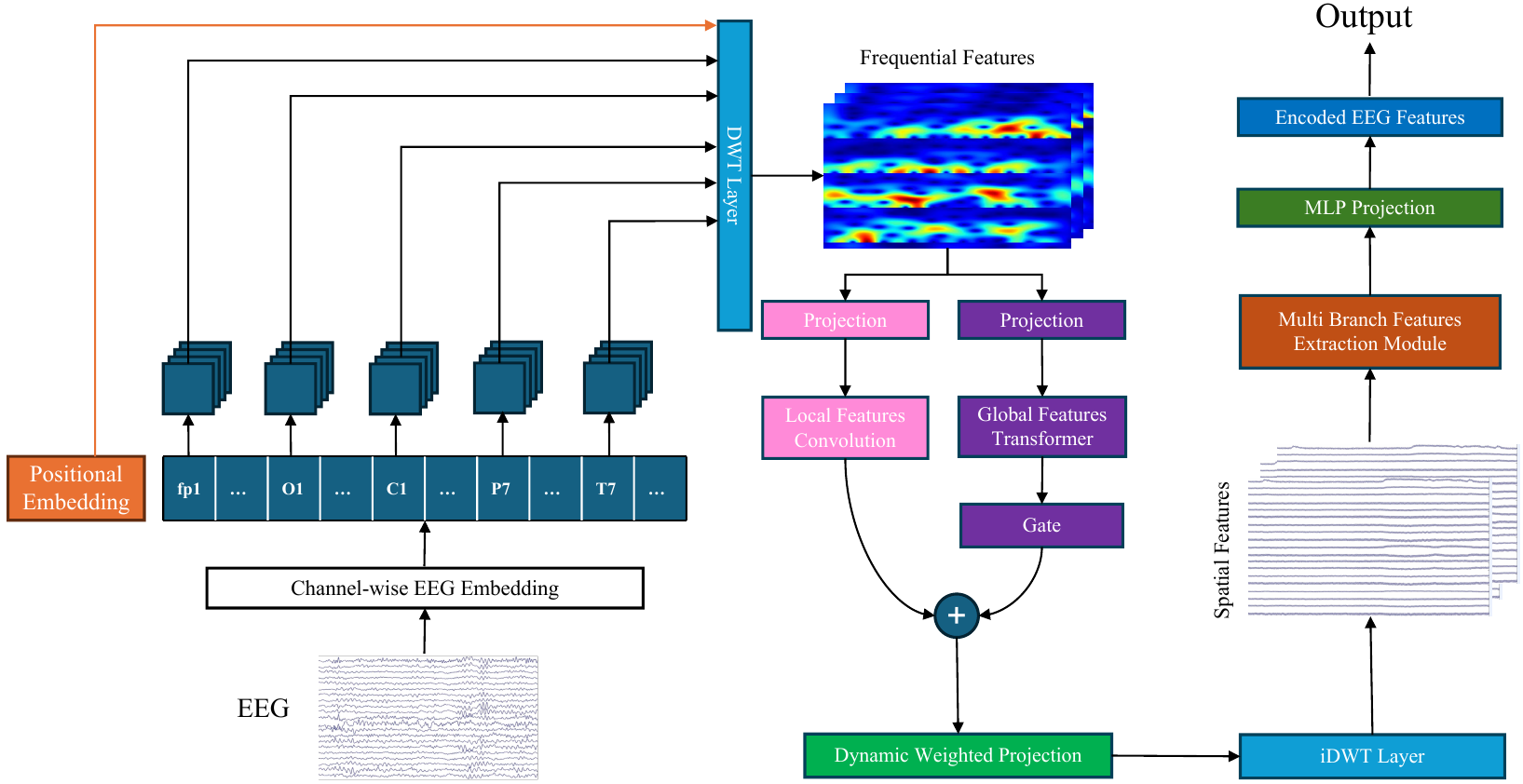}
    \caption{The overall Structure of the encoder. We employed several well-designed modules including DWT block, gated attention mechanisms and a feature-fusion modules to enhance the performance of EEG embedding generation. }
    \label{figure3}
\end{figure}

Fig.~\ref{figure2} depicts the overall flowchart of the proposed model.
In this study, the proposed model comprises three main parts: EEG embedding with a well-designed EEG encoder seen in Fig.~\ref{figure3}, the part of downstream task (image reconstruction and EEG classification), and the semantic evaluation. 
Within the EEG embedding part, several well-designed modules are employed, including DWT block, gated attention mechanisms, and a feature-fusion module.
During network training, to achieve optimal performance in terms of feature alignment and category guidance, three distinct losses are combined: the CLIP loss, the Mean Square Error (MSE) loss, and the category-aware clustering loss.
In the image reconstruction of the downstream task branch, the approach adopted is similar to that of the ATM-S method ~\cite{li2024visual}.

\subsection{DWT Block}
To effectively capture both the temporal- and frequency- domain characteristics of EEG, we propose a DWT-based feature extraction module utilizing the PyTorch Wavelet package~\cite{cotter2020uses}. 
In this block, the EEG embedding input can be written as $E \in R^{B\times C\times T}$, where $B$ represents the batch size, $C$ denotes the number of channels and $T$ is the length of the signal.
Using DWT module, the model can synthesize both frequency and spatial features at different stages, enabling the model to learn more complex features.

\subsubsection{Channel-wise DWT Decomposition}
Given the time-series characteristics and high noise levels inherent in EEG, in this study, the Daubechies-1 (db1) wavelet known as the Haar wavelet, with a single-level decomposition is used for the implementation of our DWT module.
The rationale behind this choice lies in the optimal localization properties of the db1 wavelet within the time domain, demonstrating remarkable effectiveness in capturing abrupt signal changes~\cite{ocak2009automatic}.
Furthermore, single-level decomposition strikes an appropriate balance between frequency resolution and computational efficiency, which can balance the accuracy and computational demands for EEG processing.
First, we apply one-dimensional (1D) DWT decomposition to each EEG channel independently:

\begin{equation}
   \begin{aligned}
        [cA_1^c, cD_1^c] = \text{DWT}_{1D} (E_{\{b,c,:\}}) \\
        s.t. ~~~\forall c \in \{1, ..., C\}, b \in \{1, ..., B\},
   \end{aligned}
\end{equation}
where $E_{\{b,c,:\}}$ denotes the temporal sequence for batch $b$ and channel $c$. 
$cA_1^c \in {R}^{T/2}$ denotes the approximation coefficients of the low-frequency components. which can capture the overall trends and low-frequency patterns of the EEG.
Concurrently, $cD_1^c \in {R}^{T/2}$ represents the detail coefficients of the high-frequency components, which retain the rapid changes and high-frequency characteristics of the signals.

\subsubsection{Inverse DWT Module}
We employ a 1D inverse DWT module (iDWT) to map the proposed EEG features back to the temporal domain according to the following equation:

\begin{equation}
F_{reconstructed} = \text{iDWT}_{1D}([cA_1^c, cD_1^c], \psi), 
\end{equation}
where $\psi$ represents the wavelet reconstruction parameters. 
This reconstruction process preserves not only the learned frequency-domain patterns but also temporal coherence.
To integrate both the original temporal features and the optimized ones, we propose a feature fusion module, expressed mathematically as:
\begin{equation}
F_{fused} = Conv(W_f[F; F_{reconstructed}] + b_f), 
\end{equation}
where $W_f$ and $b_f$ are learnable parameters. $F$ represents the proposed features derived from $[cA_1^c, cD_1^c]$, and $Conv$ denotes the convolution operations.

\subsection{Gated Attention Mechanism}
To process EEG embeddings from DWT modules while maintaining channel characteristics, we propose a gate-enhanced multi-head attention mechanism. This approach integrates the traditional gating mechanism within the attention of the transformer for selective feature processing. 
Our implementation features an adaptive gating integration mechanism for EEG embeddings, bridging local convolutional and global transformer features.
The core mechanism consists of temporal pooling, channel projection, and activation. Temporal pooling reduces the temporal information per channel. 
Channel-wise convolutions project values to learn weighting patterns, which undergo sigmoid activation to produce normalized gating coefficients, balancing local and global features.

\textbf{Local Feature Integration:}
The local feature extraction process begins with channel-wise temporal convolution, which processes each EEG channel independently, thus preserving their temporal relationships and individual patterns. 
Subsequently, we applied a cross-channel spatial convolution, integrating information from neighboring electrodes to effectively capture the spatial dependencies within the EEG structure. The integration of cross-channel features is realized through a comprehensive spatial processing pipeline that operates on wavelet-transformed signals. 
By applying two-dimensional convolution operations across the channel dimension of the wavelet coefficients, we effectively capture the intricate spatial relationships among different EEG channels within the frequency domain. 
This spatial integration process concatenates the wavelet coefficients from all channels before applying a spatial convolution kernel, enabling the model to learn meaningful cross-channel patterns. The convolution output then goes through batch normalization and non-linear activation, yielding a refined feature representation that simultaneously maintains the channel-specific frequency characteristics from the wavelet transform and the spatial relationships between channels. 

\textbf{Global Frequency Domain Attention:}
For global feature processing, we utilize a transformer-based encoder that employs multi-head self-attention mechanisms to capture long-range dependencies throughout the entire sequence length while preserving the channel-specific information structure. 
The integrated spatiotemporal features are then processed by a specialized channel-wise attention mechanism module. By applying a dedicated attention operation to each channel's spatiotemporal components, this module learns to selectively emphasize the most relevant patterns while simultaneously maintaining the unique characteristics of individual EEG channels. The use of multi-head attention assigns adaptive weights to different components based on their importance, enabling the model to focus on discriminative spectral features essential for downstream tasks while suppressing less informative frequency content. 

\textbf{Dynamic Weighted Projection:}
The integration process constructs a dynamic weighted combination model, where the local convolution features and the global transformer features are balanced based on the learned channel-specific weights.  
This method allows the model to automatically adjust the contribution ratio of different feature types according to the characteristics of the input signal and the specific requirements of each channel.  
As intelligent selectors, gate mechanisms excel in EEG processing by learning to balance local temporal patterns with global dependencies. 
The combination of local patterns and global dependencies has proven particularly valuable in different applications, including EEG processing~\cite{song2022eeg,lai2023interformer}, since the functional role of different channels, determined by their spatial location and measured brain activity, often requires different processing strategies. 
Therefore, the relationship between local patterns and global signal dependence can vary considerably depending on different regions of the brain and cognitive states~\cite{song2022eeg,du2023decoding}.  
The adaptive specialty of our proposed gate mechanism ensures that the model can handle these changes while preserving the integrity of the signal.
In addition, the combination of gating mechanisms with multi-head attention mechanisms offers a robust framework for EEG processing, enabling the model to concentrate on the most relevant patterns while maintaining computational efficiency.

\subsection{Multi-Branch Feature Extraction Module}
Due to the proposed multi-branch architecture, we effectively extract and fuse temporal and spatial features from EEG via parallel processing streams, capturing both time-domain characteristics and inter-channel relationships present in EEG. The temporal branch focuses on extracting time-domain characteristics from EEG by employing a series of well-designed convolution operations based on ShallowConvNet~\cite{schirrmeister2017deep}. 
First, a temporal convolution is applied across the time dimension while processing each channel independently, thereby capturing local temporal patterns during the task. Subsequently, the temporal features undergo dimensionality reduction through average pooling, and then a point-wise convolution is performed, which adjusts the feature representation while maintaining temporal relationships. 

The spatial branch is designed to capture inter-channel relationships and spatial patterns across the EEG electrode array. We implement this by initially performing a spatial convolution across the channel dimension, enabling the network to learn local spatial patterns between neighboring electrodes.
This is followed by depth-wise separable convolutions, which efficiently expand the receptive field while maintaining computational efficiency. 

The fusion mechanism employs an attention-based approach to adaptively combine temporal and spatial features. After concatenating features from both branches, we apply a channel attention mechanism that generates dynamic weights for each feature channel. The attention module consists of channel-wise average pooling followed by two convolution layers with a bottleneck structure, producing attention weights through a sigmoid activation.
The final stage of our architecture integrates the attention-weighted features through a fusion branch. A point-wise convolution combines the weighted features, followed by batch normalization and nonlinear activation. The output then undergoes a final adaptive pooling operation to ensure consistent dimensionality. 

\subsection{Loss Function}
In our approach, we introduce a dual-loss mechanism that integrates three complementary components: CLIP loss and MSE loss for general feature alignment, and a category-aware clustering loss for enhanced clustering.
By employing label-free category-aware clustering loss, we endow the model with potential zero-shot discrimination capabilities, particularly relevant in cases where the label space may differ between the training and testing phases, as observed in our dataset where the label categories of the training and testing sets are inconsistent.
During network training, these two types of losses are combined using different weights as hyperparameters. We calculate the CLIP losses, including the loss between EEG features and image features and the loss between EEG features and text features, to ensure that the EEG features can reflect relevant information. Subsequently, we calculate the category-aware clustering loss for the classification results.
The CLIP loss, $\mathcal{L}_{align}$ and the MSE loss, $\mathcal{L}_{MSE}$, between image features, $F_{I}$, and EEG features, $F_{E}$, can be written as follow:
\begin{equation}
    \mathcal{L}_{align}(F_{I}^k, F_{E}^k) = \frac{1}{2}(\mathcal{L}_{I\rightarrow E}(F_{I}^k, F_{E}^k) + \mathcal{L}_{E\rightarrow I}(F_{I}^k, F_{E}^k)),
\end{equation}
where $k$ is the index of training sample.

\subsubsection{Category-aware Clustering Loss}
A key innovation of our approach lies in implementing contrastive learning without relying on absolute class labels through class-aware clustering loss. We dynamically construct relationships within each batch by examining the relative similarities among samples. Instead of depending on predefined classes, we adaptively create positive and negative pairs based on the data in the current batch, enabling the model to learn feature representations through comparison. This similarity-based clustering operation is independent of the total number of classes in the dataset, making it particularly effective in scenarios where class distributions may vary between training and testing phases. 
The category-aware clustering loss can be formulated as follow:

\begin{equation}
    \mathcal{L}_{C}^k = \sum_{i= 1 \atop i \neq j}^J max(0, Sim(Q_j^k,C_i) - M) - log(Sim(Q_j^k,C_j) + \epsilon).
\end{equation}

In the formulas above, by element-wise multiplication, the $Sim$ refers to the similarity between the feature, $Q_j^k$, which belongs to the $k^{th}$ sample of the $j^{th}$ class, and the center of class $j$, $C_{j}$. $J$ is the amount of categories in the current batch. $\epsilon$ is used as a constant to make sure the numerical stability during calculating. And the constant $M$ is used as the threshold. When the dissimilarity achieves $M$, the loss assumes that these two samples could be completely separated. 
The final loss function can be written as follow:
\begin{equation}
    \mathcal{L}_{loss} = \sum_{k=1}^K \lambda_1\mathcal{L}_{align}(F_{I}^k, F_{E}^k) + \lambda_2\mathcal{L}_{MSE}(F_{I}^k, F_{E}^k) + \lambda_3\mathcal{L}_{C}^k,
\end{equation}
where K is the number of samples in the current batch. $\lambda_1$, $\lambda_2$ and $\lambda_3$ are the weighting parameters to balance these three loss functions.

\subsection{Dataset}
We conducted our experiments using the THINGS-EEG~\cite{gifford2022large} dataset, which is a large-scale collection of EEG recordings designed to capture neural signals to visual stimuli. 
The THINGS-EEG dataset was developed as part of the THINGS initiative\footnote{https://things-initiative.org/}, aiming at exploring the neural representation of object concepts. It comprises EEG recorded from participants who were exposed to a diverse set of real-world object images. These images were meticulously selected from the broader THINGS dataset~\cite{hebart2019things}, which encompasses over 26,000 object concepts spanning a wide semantic and perceptual range. 
Consequently, the visual stimuli in THINGS-EEG are diverse and representative, facilitating the study of rich and varied neural responses.
The EEG data were recorded using a 64-channel EASYCAP, with electrodes arranged according to the standard 10–10 system, and a Brainvision actiCHamp amplifier. The sampling rate was set at 1000 Hz, with online filtering applied between 0.1$Hz$ and 100$Hz$, and referencing to the Fz electrode. 
MNE python package was applied to conduct the preprocessing.~\cite{GramfortEtAl2013a}. The continuous EEG data were segmented into trials from 200$ms$ before stimulus onset to 800$ms$ after stimulus onset. Baseline correction was implemented by subtracting the mean pre-stimulus interval for each trial and channel. Subsequently, the data were downsampled to 100$Hz$. Trials containing target stimuli were excluded from our experiment. For our experiments, we employed the same training and testing datasets from the THINGS-EEG dataset as those used in previous research.

\subsection{Evaluation}
In our experiments, we primarily used the classification accuracy including Top-1 accuracy and Top-5 accuracy to evaluate the capability of EEG encoder. 
For the reconstruction of visual stimuli, traditional quantitative metrics, such as MSE and SSIM, are not appropriate. This is because we only need to reconstruct images with consistent semantics and do not require pixel-level reconstruction recovery. 
Therefore, to assess the semantic information of the generated images, we creatively propose a semantic-based score as an evaluation metric for the quality of the generated images, based on the image classification model pre-trained on ImageNet, which, to the best of our knowledge, is the first time to be proposed.
Specifically, we utilize the pre-trained weights of the ConvNext model from Meta~\cite{liu2022convnet} to classify the generated images. 
For any generated image, we adopt the following rules:

\begin{enumerate}
\item If the result of Top-1 from the ConvNext model contains the same label, or the THINGS-EEG labels are included in the ImageNet labels, the score of the image is set to 1.
\item If the results of Top-5 from the ConvNext model contain the same label, the score of the image is set as the sum of the probabilities of the classes that contain the same label.
\item If the results of Top-5 from the ConvNext model do not contain the exact same label, we will use WordNet~\cite{miller1995wordnet} to obtain the label's classes.
\end{enumerate}
Then we calculate the score using the following formula:
\begin{equation}
Score = \sum P_i * Sim(label, WordNet_{label})
\end{equation}
Here, $P_i$ represents the probability of each class obtained from the pre-trained ConvNext model, and $Sim$ is the Wu-Palmer Similarity~\cite{wu1994verb}.

Here, we carefully selected several representative methods, as comparison methods, including ATM-S, ATM-E~\cite{li2024visual}, NERV~\cite{chen2024necomimi}, NICE~\cite{song2023decoding}, EIT-ResNet~\cite{zheng2024eit,he2016deep}. 

Also, in the following part, we compared the generated images in the same subject, which is the subject "sub-08" to assess whether the generated images are able to reflect the semantic information successfully.

\subsection{Implementation Details}
In this work, we well-trained the proposed model with a maximum of 40 epochs, a learning rate of $3\times10^{\text{-}4}$, and a batch size of 64. Early stopping was applied, and training was halted if the top-1 classification accuracy did not improve for 10 consecutive epochs. To complement the evaluation, we also recorded the accuracy of the top 5 classification to capture the five most probable class predictions, which were further applied to analyze semantic consistency. In the whole training process, we also applied pre-trained models in different parts to reduce the computation cost and increase the generality. First, we applied the public CLIP weights to generate image embedding from the testing and training dataset of THINGS-EEG. Then, we applied the pre-trained diffusion models including SDXL-Turbo to generate images~\cite{podell2023sdxl,sauer2025adversarial,ye2023ip}. Finally, we used the pre-trained ConvNext trained with ImageNet~\cite{deng2009imagenet} and wordNet~\cite{miller1995wordnet} to calculate the semantic scores based on the classification results.

\section{Experimental Results}
\begin{figure}[htbp]
\centering
    \includegraphics[width=0.48\textwidth]{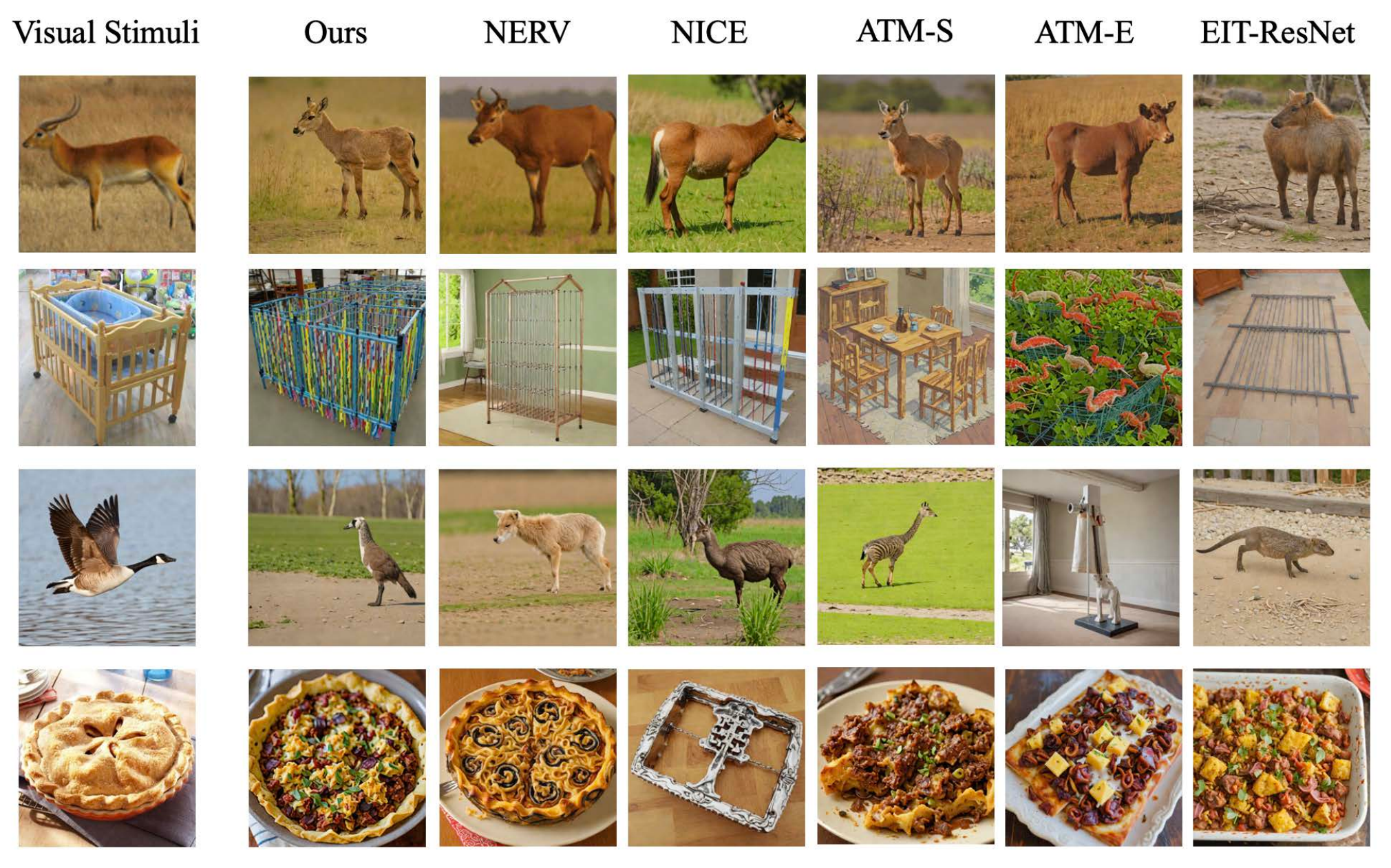}
    \caption{We present several examples from subject-08. The left column shows the visual stimuli corresponding to the EEG signals, while the subsequent columns display the reconstructed results of different methods respectively.
    These results exhibit that our model is able to learning semantic features from EEG. }
    \label{visual}
\end{figure}

\subsection{Visual Inspection of Image Reconstruction}
To evaluate the ability of our framework to reconstruct high-quality images from EEG, we conducted relevant image-generation tasks on the THINGS-EEG dataset using all relevant model reconstructions.
Fig.~\ref{visual} exhibits the representative reconstructed images from subject-08. 
The first column presents the visual stimuli corresponding to the EEG signals. In terms of the semantic quality of the generated images, significant differences exist among the results produced by different methods shown in the subsequent columns. The second column displays the images generated by our method. 
From Fig.~\ref{visual}, we can see that When faced with the visual stimulus of "antelope", both our method and ATM-S retain, to a certain extent, the key features related to "antelope". Although pixel-level precise reconstruction may not be achieved, the overall outlines and main structures are highly recognizable. For example, the shape of the antelope can be roughly outlined. However, the images reconstructed by other methods resemble either cows or horses.
For another instance, considering the visual stimulus of "bird" in the third row, only our method can, to some degree, make the generated image recognizable as a bird. Other methods fail to do so. In particular, the image reconstructed by ATM-E is not an animal-related image at all, showing a huge semantic discrepancy.

\begin{table}[]
\centering
\caption{The results of semantic scores of different models in subject-08.}
\label{Semantic_score}
\resizebox{0.3\textwidth}{!}{
\begin{tabular}{ccc}
\toprule
\textbf{Method} &\textbf{Mean} &\textbf{Standard deviation}
\\ \midrule
\textbf{Ours}	&\color{red}\textbf{0.383} &0.182\\
\textbf{NERV}	&0.376 &0.198\\
\textbf{NICE} &0.364 &0.167\\
\textbf{ATM-S} &0.368 &0.176\\
\textbf{ATM-E} &0.326 &0.154\\
\textbf{EIT-ResNet} &0.343 &0.147\\
\bottomrule
\end{tabular}}
\end{table}

\subsection{Quantitative Semantic Analysis}

\begin{figure}[htbp]
\centering
    \includegraphics[width=0.48\textwidth]{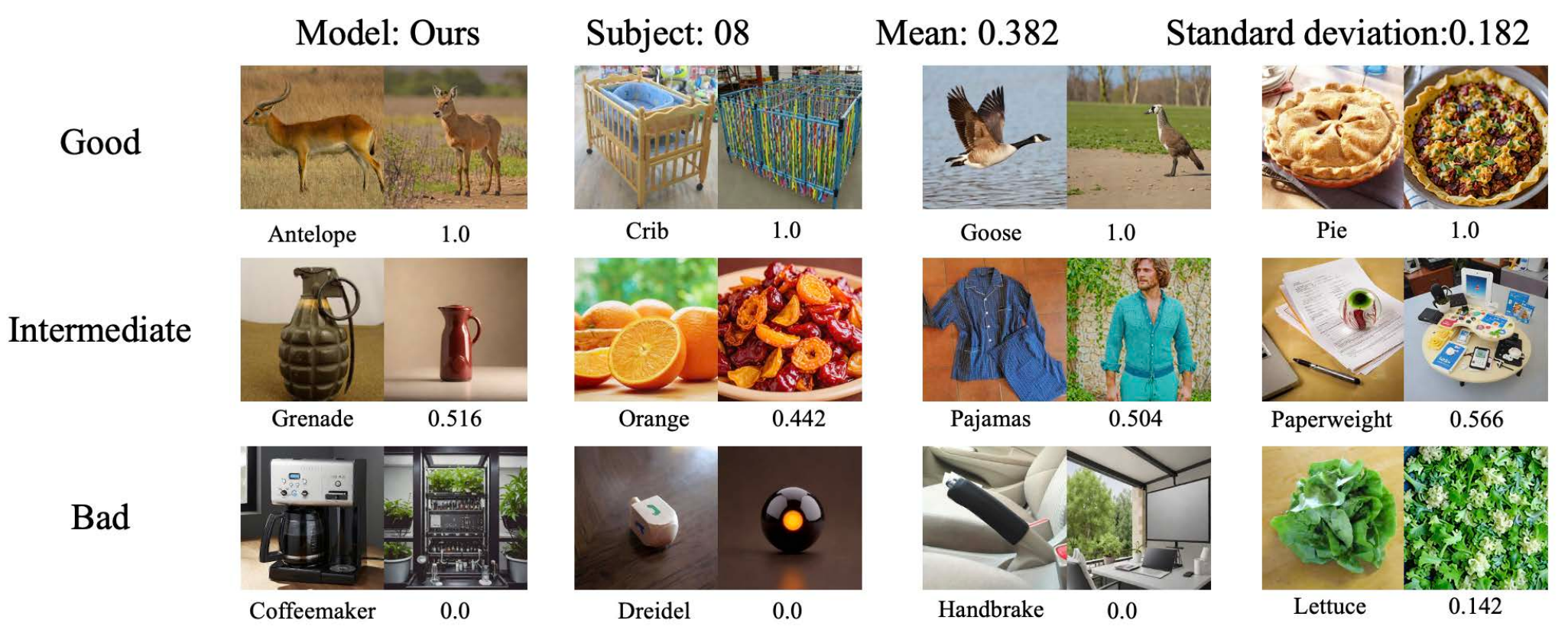}
    \caption{Different semantic scores of generated images in subject-08. For each group of two images, the left one refers to the visual stimuli, and the right one refers to the representative generated result. The number below the right image refers to the semantic score. We refer the good images as the score over (mean + std), the bad images as the score below (mean - std), the left images as intermediate. }
    \label{visual_only}
\end{figure}

From Fig.~\ref{visual}, it can be observed that although our method cannot reconstruct visual stimuli at the pixel level, it is capable of generating images with similar semantics. 
To quantitatively compare the semantic similarity of images generated by different methods, in this work, we propose a quantitative semantic-based score. 
We categorize the scores of all generated images into three types based on the mean and standard deviation. For scores above (mean + std), the images are classified as good. For scores between (mean + std) and (mean - std), the images are classified as intermediate. For scores below (mean - std), the images are classified as bad.
From Fig.~\ref{visual_only}, we can see that, to some extent, The higher the score, the closer the semantics of the reconstructed image is to the visual stimuli corresponding to the EEG signals.
By comparing the magnitudes of these scores, we can, to a certain extent, conduct a horizontal comparison of the quality of images reconstructed by different methods, seen in Table~\ref{Semantic_score}.

\subsection{Zero-shot EEG Classification}

In the process of our EEG reconstruction of visual stimuli, a powerful EEG feature extractor is required to better extract corresponding semantic information from EEG.
Evaluation of the ability of EEG feature extraction to extract relevant semantic information from EEG can be used to assess the ability of each model to reconstruct EEG visual stimuli.
Therefore, we conducted a zero-shot classification task using the THINGS-EEG testing dataset. 
By aligning the EEG features with image features through the CLIP loss, we simultaneously optimize the EEG features via category-aware clustering loss, enabling the separation of different representations in the high-dimensional space while clustering similar ones. Subsequently, these features are processed by a lightweight Multi-Layer Perceptron (MLP) classifier.

The results, as presented in Table~\ref{Table_1}, demonstrate that our model achieves superior performance compared to other baseline methods. 
To be specific, although our method does not achieve the best Top-1 and Top-5 accuracy across all subjects, overall, it demonstrates the best performance, which can be clearly observed from the average precision. This suggests that our method is more stable and effective in handling the task in general, rather than excelling in only a few specific cases.

\begin{table*}[htbp]
\centering
\scriptsize 
\setlength{\tabcolsep}{2pt} 
\renewcommand{\arraystretch}{1.2} 

\caption{The overall accuracy for the classification task.
We implemented the models from previous papers including NERV, ATM-S, ATM-E, \textit{etc.} 
We calculated the subject-wise top-1 and top-5 accuracy of classification results. 
We found that compared with other state-of-the-art methods such as NERV and ATM-S, our proposed model still advance in both mean Top-1 and top-5 accuracy.}

\label{Table_1}

\begin{tabular}{ccccccccccccccccccccccc}
\toprule
\textbf{Method}  & \multicolumn{2}{c}{\textbf{Sub-01}} & \multicolumn{2}{c}{\textbf{Sub-02}} & \multicolumn{2}{c}{\textbf{Sub-03}} & \multicolumn{2}{c}{\textbf{Sub-04}} & \multicolumn{2}{c}{\textbf{Sub-05}} & \multicolumn{2}{c}{\textbf{Sub-06}} & \multicolumn{2}{c}{\textbf{Sub-07}} & \multicolumn{2}{c}{\textbf{Sub-08}} & \multicolumn{2}{c}{\textbf{Sub-09}} & \multicolumn{2}{c}{\textbf{Sub-10}} & \multicolumn{2}{c}{\textbf{mean}}\\ 

                 & \textbf{Top-1} & \textbf{Top-5} & \textbf{Top-1} & \textbf{Top-5} & \textbf{Top-1} & \textbf{Top-5} & \textbf{Top-1} & \textbf{Top-5} & \textbf{Top-1} & \textbf{Top-5} & \textbf{Top-1} & \textbf{Top-5} & \textbf{Top-1} & \textbf{Top-5} & \textbf{Top-1} & \textbf{Top-5} & \textbf{Top-1} & \textbf{Top-5} & \textbf{Top-1} & \textbf{Top-5} &\textbf{Top-1} & \textbf{Top-5}\\ 

\midrule
\textbf{Ours}             & \color{red}\textbf{33.0} & \color{red}\textbf{58.5}  & \color{red}\textbf{28.0} & \color{red}\textbf{56.5}  & 33.5 & \color{red}\textbf{61.0}  & \color{red}\textbf{36.0} & \color{red}\textbf{68.0}  & \color{red}\textbf{26.0} & 48.0  & 30.5 & 62.5  & \color{red}\textbf{34.0} & \color{red}\textbf{62.5}  & \color{red}\textbf{43.0} & 73.5  & \color{red}\textbf{31.5} & 58.5  & \color{red}\textbf{38.5} & 69.0  &\color{red}\textbf{33.4}	&\color{red}\textbf{61.8}\\ 
NERV           &25.5	&57	&26	&57	&31.5	&60.5           &30.0	&58.0	&23.0	&\color{red}\textbf{50.0}	&\color{red}\textbf{32.0}	&55.5	&31.0	&61.0	&40.5	&70.5	&30.5	&60.5	&34.0	&66.0     &30.4 &59.64      \\ 
NICE &25.0	&49.0	&17.5	&45.5	&30.5	&55.5	&33.0	&60.5	&14.0	&37.5	&28.5	&54.0	&24.5	&52.5	&39.0	&70.5	&23.5	&45.5	&30.0	&62.0	&26.55	&53.25 \\
ATM-S    &24.0	&56.5	&24.5	&53.0	&\color{red}\textbf{34.5}	&60.0	&35.0	&65.0	&20.5	&48.5	&31.0	&\color{red}\textbf{65.5}	&31.5	&\color{red}\textbf{62.5}	&40.5	&\color{red}\textbf{75.0} &30.0	&\color{red}\textbf{59.0}	&\color{red}\textbf{38.5}	&\color{red}\textbf{70.0}	&31.0	&61.5 \\ 
ATM-E	&19.0	&49.5	&18.5	&40.0	&28.5	&59.5	&31.0	&56.0	&17.5	&39	&23.5	&52.0	&25.5	&53.0	&30.5	&64.0	&24.5	&49.0	&32.0	&60.5 &25.05	&52.25\\
EIT-ResNet &16.5 	&30.5 	&11.5 	&26.5 	&13.0 	&38.0 	&14.0 	&32.0 	&8.0 	&24.5 	&15.5 	&39.0 	&16.5 	&40.0 	&16.0 	&39.5 	&12.5 	&26.0 	&14.5 	&43.5 	&13.8 	&34.0 \\
\bottomrule
\end{tabular}

\end{table*}

\begin{table*}[]
\centering
\caption{The results for the ablation study in classification task. }
\label{ablation}
\resizebox{\textwidth}{!}{
\begin{tabular}{@{}cccccccccccccccccc@{}}
\toprule
& \textbf{$\mathcal{L}_{align}$}        & \textbf{$\mathcal{L}_{MSE}$}              & \textbf{$\mathcal{L}_{C}$}    & \textbf{DWT}              & \textbf{Local branch}            & \textbf{Global branch}           & \textbf{Sub-01} & \textbf{Sub-02} & \textbf{Sub-03} & \textbf{Sub-04} & \textbf{Sub-05} & \textbf{Sub-06} & \textbf{Sub-07} & \textbf{Sub-08} & \textbf{Sub-09} & \textbf{Sub-10} & \textbf{mean} \\ \midrule
\multirow{6}{*}{\textbf{\begin{tabular}[c]{@{}c@{}}Gated attention \\ in global branch\end{tabular}}} & \ding{55}     & \Checkmark & \Checkmark & \Checkmark & \Checkmark & \Checkmark &1.0                 &0.5                 &1.0                   &0.5              &1.0                 &0.5                 &1.0                 &1.0                 &0.5                 &1.0                 &0.8              \\ \cline{2-18}
& \Checkmark & \ding{55}     & \Checkmark & \Checkmark & \Checkmark & \Checkmark & 29.5            & 26.0            & 31.0            & 34.0            & 19.0            & 28.0            & 32.0            & 38.5            & 28.5            & 35.0            & 30.15         \\ \cline{2-18}
& \Checkmark & \Checkmark & \ding{55}     & \Checkmark & \Checkmark & \Checkmark & 30.0            & 27.0            & 31.5            & 37.5            & 22.5            & 31.5            & 34.5            & 42.5            & 31.5            & 39.5            & 32.8          \\ \cline{2-18}
& \Checkmark & \Checkmark & \Checkmark & \ding{55}     & \Checkmark & \Checkmark & 20.5            & 25.0            & 30.5            & 29.0            & 20.0            & 25.5            & 24.5            & 42.5            & 25.5            & 30.5            & 27.35         \\ \cline{2-18}
& \Checkmark & \Checkmark & \Checkmark & \Checkmark & \ding{55}     & \Checkmark & 23.5            & 20.5            & 28.5            & 24.0            & 17.0            & 26.0            & 24.5            & 34.5            & 23.0            & 26.0            & 24.75         \\ \cline{2-18}
& \Checkmark & \Checkmark & \Checkmark & \Checkmark & \Checkmark & \ding{55}      & 21.0            & 26.5            & 33.0            & 28.0            & 20.5            & 27.5            & 24.5            & 40.0            & 27.0            & 27.0            & 27.5          \\ \hline
\textbf{\begin{tabular}[c]{@{}c@{}}Gated attention\\  in local branch\end{tabular}}                   & \Checkmark & \Checkmark & \Checkmark & \Checkmark & \Checkmark & \Checkmark & 20.5            & 26.5            & 32.0            & 30.5            & 21.0            & 28.5            & 26.5            & 41.5            & 23.0            & 27.5            & 27.75         \\ \hline
\multicolumn{7}{c}{\textbf{Ours}}                                                                                                                           & \textbf{33.0}   & \textbf{28.0}   & \textbf{33.5}   & \textbf{36.0}   & \textbf{26.0}   & \textbf{30.5}   & \textbf{34.0}   & \textbf{43.0}   & \textbf{31.5}   & \textbf{38.5}   & \textbf{33.4} \\ \bottomrule
\end{tabular}
}
\end{table*}

\subsection{Ablation Study}

To explore the impact of different components on the final performance of the model, we conducted ablation studies from several perspectives. First, based on our model, we removed the feature alignment loss. As shown in Table~\ref{ablation}, we found that $\mathcal{L}_{align}$ is the most crucial loss for training our model. Without the assistance of $\mathcal{L}_{align}$, the accuracy of EEG classification becomes extremely low, indicating that the model cannot fit the data under such circumstances.
Furthermore, we removed the DWT module, the local branch, and the global branch separately to train the classification task for each subject. The accuracy of the classification task decreased significantly across all subjects. A similar phenomenon also occurred in another ablation experiment: when we changed the gated attention from the global branch to the local branch.
Regarding the other two loss functions $\mathcal{L}_{MSE}$ and $\mathcal{L}_{C}$ during the training process, we found that removing either one of them led to a slight decline in the performance of our model.

\section{Discussion}

EEG, as an indispensable tool in BCI, enables non-invasive acquisition of relevant brain information. Through analysis, EEG can be utilized to understand brain-related activities and guide clinical treatment and assisted living for relevant individuals.
Among them, the reconstruction of relevant visual stimuli from EEG can make EEG more tangible, which is conducive to connecting individuals with the real world.

In this work, we present a novel framework for EEG-to-image reconstruction, which enhances the extraction of meaningful features from EEG.
To be specific, the proposed model integrates advanced components such as the DWT block and gated attention mechanisms, as well as a novel mixed loss function that combines CLIP loss for feature alignment with a label-free category-aware clustering loss, aiming to improve classification accuracy and semantic alignment. 
Furthermore, we innovatively utilize WordNet-based classification accuracy and semantic similarity measures and creatively proposed a semantic-based score to objectively evaluate the semantic information of generated images, addressing limitations in existing evaluation methods and providing a scalable and quantitative way to assess the semantic consistency of reconstructed images.

Compared with previous results, EIT~\cite{zheng2024eit} applied mature deep learning tools such as ResNet as the EEG feature extractor for image reconstruction. However, despite its advance in efficiency and computation cost, the nature of such computer vision models may not be suitable for processing complex time-series data, especially for EEG data and the complicated downstream tasks such as high-level EEG feature extraction and feature alignment with images.  
Compared to NERV~\cite{chen2024necomimi} and ATM-S~\cite{li2024visual}, our model successfully introduced the DWT module as a frequency-domain feature extractor.
Also, the application of category-aware clustering loss leads to further performance improvements.

Based on our analysis, there is a strong correlation between the classification accuracy and the semantic score of the generated images. Specifically, EEG that are correctly classified by the model typically result in images with higher semantic similarity scores. In contrast, EEG that are not correctly classified usually produce images with significantly lower semantic similarity scores. This finding emphasizes the importance of accurate classification in maintaining semantic fidelity during the EEG-to-image conversion process.
Challenges found when generating images for specific categories: Some categories completely fail to generate semantically meaningful images. We have identified several potential reasons for this limitation: (1) Ambiguous or incorrect labels: Categories with unclear or incorrect labels, such as "bator4", pose significant challenges. Even when converted into semantic vectors through CLIP, these labels lack meaningful semantic relationships, leading to image generation failures. (2) Diffusion model constraints: In this paper, the directly used pre-trained diffusion model for reconstructing visual stimuli may lack corresponding visual representations for certain categories, limiting its ability to generate relevant images.

Despite numerous challenges, this study highlights the potential of EEG-based image generation as a research frontier. Future work could explore the use of interpretable models or traditional EEG analysis methods to extract more robust features, thereby enhancing the interpretability of the process. Moreover, improving the consistency between EEG features and semantic representations through more sophisticated architectures or loss functions could further advance the field. Developing customized pre-training strategies for diffusion models to include a broader range of categories is another promising avenue for improving image generation results.

In conclusion, our findings demonstrate the feasibility and value of EEG-based image generation, offering insights into the semantic encoding of neural signals and paving the way for more interpretable and accessible brain-computer interface technologies.

\section*{Ethical Statement}

There are no ethical issues.

\section*{Acknowledgments}

This work was partially supported by the Beijing Natural Science Foundation (L245015), the National Key R\&D Program Program of China (2023YFC2706400), the Chongqing Natural Science Foundation (CSTB2024NSCQ-MSX0451), and the Medical Innovation Program(MIP) in Kyoto University. The authors also would like to show a great appreciation towards Dr.Yoshifumi Mori, who now works in Netherlands Institute for Neuroscience, for his valuable suggestions and insights in psychiatric diseases.

\bibliographystyle{named}
\bibliography{ijcai25}

\end{document}